\documentclass{article}

\usepackage{arxiv}

\usepackage[utf8]{inputenc} 
\usepackage[T1]{fontenc}    
\usepackage{hyperref}       
\usepackage{url}            
\usepackage{booktabs}       
\usepackage{amsfonts}       
\usepackage{nicefrac}       
\usepackage{microtype}      
\usepackage{amsmath}
\usepackage{cleveref}       
\usepackage{lipsum}         
\usepackage{graphicx}
\usepackage{natbib}
\usepackage{doi}
\usepackage{authblk}

\usepackage[T1]{fontenc}
\usepackage{graphicx}
\usepackage{booktabs}
\usepackage[misc]{ifsym}

\usepackage{graphicx}
\usepackage{blindtext}
\usepackage{glossaries}
\usepackage{mathtools}
\usepackage{wrapfig}
\usepackage{algorithm}
\usepackage{wrapfig}
\usepackage[noend]{algpseudocode}
\usepackage{multirow}
\usepackage{subfig}
\usepackage[table, svgnames, dvipsnames]{xcolor}
\usepackage{makecell, cellspace, caption}
\usepackage{fancyhdr}

\newacronym{CNF}{CNF}{Conjuctive Normal Form}
\newacronym{ARM}{ARM}{Association Rule Mining}
\newacronym{NARM}{NARM}{Numerical Association Rule Mining}
\newacronym{ARL}{ARL}{Association Rule Learning}
\newacronym{RDF}{RDF}{Resource Description Framework}
\newacronym{IoT}{IoT}{Internet of Things}
\newacronym{DL}{DL}{Deep Learning}

\title{Learning Semantic Association Rules from Internet of Things Data}

\fancypagestyle{firstpage}{
    \fancyhf{}
    
    \fancyhead[C]{\footnotesize This paper has been published in Neurosymbolic Artificial Intelligence Journal (SAGE), DOI: 10.1177/29498732251377518.}
}

\newbox{\orcid}\sbox{\orcid}{\includegraphics[scale=0.06]{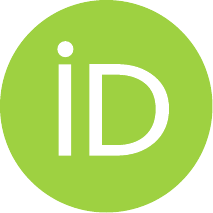}} 
\author[1]{%
	\href{https://orcid.org/0000-0003-2710-7951}{\usebox{\orcid}\hspace{1mm}Erkan Karabulut}%
}
\author[1]{%
	\href{https://orcid.org/0000-0003-0183-6910}{\usebox{\orcid}\hspace{1mm}Paul Groth}
}
\author[1]{%
	\href{https://orcid.org/0000-0001-7054-3770}{\usebox{\orcid}\hspace{1mm}Victoria Degeler}%
}
\affil[1]{University of Amsterdam, 1098 XH, North Holland, The Netherlands \protect\\ \texttt{\{e.karabulut, v.o.degeler, p.t.groth\}@uva.nl}}

\begin{document}
\maketitle

\thispagestyle{firstpage}

\begin{abstract}
Association Rule Mining (ARM) is the task of discovering commonalities in data in the form of logical implications. ARM is used in the Internet of Things (IoT) for different tasks including monitoring and decision-making. However, existing methods give limited consideration to IoT-specific requirements such as heterogeneity and volume. Furthermore, they do not utilize important static domain-specific description data about IoT systems, which is increasingly represented as knowledge graphs. In this paper, we propose a novel ARM pipeline for IoT data that utilizes both dynamic sensor data and static IoT system metadata. Furthermore, we propose an Autoencoder-based Neurosymbolic ARM method (Aerial) as part of the pipeline to address the high volume of IoT data and reduce the total number of rules that are resource-intensive to process. Aerial learns a neural representation of a given data and extracts association rules from this representation by exploiting the reconstruction (decoding) mechanism of an autoencoder. Extensive evaluations on 3 IoT datasets from 2 domains show that ARM on both static and dynamic IoT data results in more generically applicable rules while Aerial can learn a more concise set of high-quality association rules than the state-of-the-art with full coverage over the datasets.
\end{abstract}

\keywords{association rule mining, neurosymbolic AI, semantic web, autoencoder, internet of things, sensor data}

\section{Introduction}



\gls{ARM} is a common data mining task that aims to discover associations between features of a given dataset in the form of logical implications~\citep{agrawal1994fast}. In \gls{IoT} systems, ARM methods are utilized for various tasks including monitoring, decision-making, and optimization, for example, of a system's resources~\citep{sunhare2022internet}. Some IoT application domains in which ARM has been successfully utilized include agriculture~\citep{iotdataminingagriculture2021}, smart buildings~\citep{itemsets-smarthome} and energy~\citep{iotdataminingenergy2023}. However, most applications of ARM in IoT give limited considerations to characteristics of IoT data such as heterogeneity and volume~\citep{ma2013data} as they are mere adaptations of rule mining methods not specifically tailored to IoT requirements. 

IoT systems can produce or use data from diverse sources which can be categorized as static and dynamic. Static data refers to data that is not subject to frequent changes such as system models while dynamic data is subject to frequent changes, for instance, sensor data. The static part of IoT systems is increasingly represented as knowledge graphs~\citep{rhayem2020semantic,karabulut2023ontologies}, large databases of structured semantic information~\citep{kg-book}. ARM algorithms are often run on the dynamic part of IoT data, not utilizing the valuable information in knowledge graphs. In addition, ARM algorithms can generate a high number of rules as the input dimension increases~\citep{kaushik2023numerical,telikani2020survey}, which is time-consuming to process and maintain. Generating a high number of rules can be the case for large-scale IoT environments, as each sensor is treated as a different data dimension. 

To address these two issues, this paper presents two new contributions. The first contribution is a novel ARM pipeline for IoT data that combines knowledge graphs and sensor data to learn association rules with semantic properties, \textit{semantic association rules} (Section \hyperref[sec:problem-statement]{Problem Statement}), that represent IoT data as a whole (Section \hyperref[sec:pipeline]{Pipeline}). We hypothesize that semantic association rules are more generically applicable than association rules based on sensor data only, requiring fewer rules to have full data coverage. As an example, an association rule based on sensor data only looks as follows: \textit{`if sensor1 measures a value in range R, then sensor2 must measure a value in range R2'}. This rule can only be applied to \textit{sensor1} and \textit{sensor2}. In contrast, semantic association rules are more contextual as seen in the following example in the water network domain: \textit{`if a water flow sensor placed in a pipe P1 with diameter $\ge$ A1 measures a value in range R, then a water pressure sensor placed in a junction J1 connected to P1 measures a value in range R2'}. The semantic association rule is no longer about individual sensors. Instead, it describes a certain context that the sensor is placed in and therefore is more generically applicable and explainable. 

However, enriching sensor data with semantics from a knowledge graph increases input size and may result in a high number of rules. Hence, the second contribution of this paper is an Autoencoder-based~\citep{denoisingautoencoder} Neurosymbolic ARM method (Aerial) as part of the proposed pipeline that can learn a concise set of high-quality rules with full data coverage (Section \hyperref[sec:rule-extraction-autoencoder]{Rule Extraction from Autoencoders}). Aerial learns a neural representation of a given input data and then extracts association rules from the neural representation. This approach can be supplemented by and is fully compatible with other ARM variations that aim to mine a smaller subset of high-quality rules such as top-k rules mining~\citep{fournier2012mining}, and ARM with item constraints~\citep{baralis2012generalized,srikant1997mining}. An extensive set of experiments (Section \hyperref[sec:experiments]{Evaluation}) is performed and the results show that ARM on knowledge graphs and sensor data together results in more generically applicable rules with high support and data coverage in comparison to ARM on sensor data only (Section \hyperref[sec:discussion]{Discussion}). Furthermore, the results show that the proposed Aerial approach is capable of learning a concise set of high-quality rules with full coverage over the entire data.

In summary, the two contributions of this paper are: (1) a pipeline of operations to learn contextual and more generically applicable semantic association rules from IoT data compared to existing methods; and (2) an Autoencoder-based ARM approach for learning a more concise set of high-quality semantic association rules than the state-of-the-art, with full data coverage. This approach is orthogonal and can be used with other ARM variations.



\section{Related Work} \label{sec:related-work}

This section introduces the related work and background concepts.

\subsection{Association Rule Mining}

ARM is the problem of learning commonalities in data in the form of logical implications, e.g., $X \rightarrow Y$, which is read as \textit{`if X then Y'}. Initial ARM algorithms such as Apriori~\citep{agrawal1994fast} and HMine~\citep{pei2001h} focused on mining rules from categorical datasets. The initial methods needed pre-discretization for numerical data, struggled with scaling on big high-dimensional data, and produced a high number of rules that are costly to post-process. FP-Growth~\citep{han2000mining}, a widely used ARM algorithm, has many variations to tackle some of the aforementioned issues. ARM with item constraints~\citep{srikant1997mining} is an ARM variation that focuses on mining rules for the items of interest rather than all, which reduces the number of rules and execution time~\citep{baralis2012generalized}. Guided FP-Growth~\citep{guidedfpgrowth} is an FP-Growth variation for ARM with item constraints. Other variations include Parallel FP-Growth~\citep{parallelfpgrowth} and FP-Growth on GPU~\citep{gpufpgrowth} for better execution times. 

Recently, a few DL-based ARM algorithms have been proposed. Patel et al.~\citep{patel2022innovative} proposed to use Autoencoders~\citep{chen2023auto} to learn frequent patterns in a grocery dataset, however, no source code or pseudo-code was given. Berteloot et al.~\citep{berteloot2023association} also utilized Autoencoders (ARM-AE) to learn association rules directly from categorical tabular datasets. However, ARM-AE has fundamental issues while extracting association rules from an Autoencoder, which we elaborate on in Section \hyperref[sec:setting2]{Setting 2: Aerial vs state-of-the-art}.

\gls{NARM} aims to identify intervals for numerical variables to generate high-quality association rules based on specific quality criteria. Following the recent systematic literature reviews~\citep{telikani2020survey,kaushik2023numerical}, the state-of-the-art in NARM is nature-inspired optimization-based algorithms which include evolutionary, differential evolution, swarm intelligence, and physics-based approaches. They employ heuristic search processes to find association rules that optimize one or more rule quality criteria and are used for both numerical and categorical datasets~\citep{fister2018differential}. However, optimization-based ARM  methods too suffer from handling big high-dimensional data, together with other broader issues in NARM such as having a large number of rules, and explainability as also mentioned by Kaushik et al. and other works~\citep{telikani2020survey,berteloot2023association,kishore2021applications}. 

\subsection{Association Rule Mining in Internet of Things}

In IoT, both exhaustive ARM, such as Apriori and FP-Growth, and the optimization-based NARM methods are used for various tasks. \cite{earlywarnings} utilized the Apriori algorithm for big data mining in IoT in the enterprise finance domain for financial risk detection. \cite{abcruleminer} utilized an exhaustive ARM approach with item constraints on phone usage data to learn user behaviors. \cite{anovelassociation} proposed a distributed exhaustive ARM approach that can run on a wireless sensor network. \cite{fister2023time}, proposed TS-NARM, an optimization-based NARM approach, and evaluated it on a smart agriculture use case with 5 optimization-based methods. 

Sequential or temporal ARM is another ARM variant used in IoT~\citep{sequentialfuzzy}. The goal is to learn patterns between subsequent events, rather than events that happen in the same time frame, concurrent events. In this paper, we focus on mining association rules for concurrent events, rather than sequential events which is a different task.

Based on recent surveys~\citep{karabulut2023ontologies,listl2024knowledge}, semantic web technologies such as ontologies~\citep{gruber1993ontology} and knowledge graphs~\citep{kg-book} have been used for knowledge representation in IoT, providing valuable knowledge related to IoT systems and its components. Naive SemRL~\citep{karabulut2023semantic} is the only ARM method that utilizes semantics when learning rules from pre-discretized sensor data. It is based on FP-Growth, however, the paper does not provide a complete evaluation. We adopt a similar semantic enrichment approach but develop a completely new DL-based pipeline, and provide an extensive evaluation. 

\begin{table}[t]
    \centering
    \caption{Input notation, explanations, and examples from water networks domain.}
    \label{tab:input-notation}
    \begin{tabular}{p{0.07\textwidth}p{0.43\textwidth}p{0.43\textwidth}}
        \toprule
        \textbf{Notation} & \textbf{Explanation} & \textbf{Example} \\ 
        \midrule
        C & Classes in an Ontology/Data schema & Pipe, Junction \\
        R, r & Relations (R) in between the classes (C) mapped with (r) & (Pipe)\_connectedTo\_(Junction) \\
        A, a & Properties for the classes and relations & (Junction).elevation: elevation property of the class Junction \\
        V & Node IDs in the knowledge graph & P1, J2 \\
        E, e & IDs of the edges (E) in between nodes (V) in the knowledge graph mapped with (e) & (P1)\_(e1)\_(J2), P1 and J2 are node IDs, e1 is an edge ID \\
        L, l & Labels for the nodes (V) and edges (E) in the knowledge graph mapped with (l) & (P1:Pipe)\_(e1:connected\_to) \_(J2:Junction) \\
        P, U, p & Property (P) and value (U) pairs for nodes and edges mapped with (p) & (P1:Pipe).elevation=v1, the elevation of pipe P1 is v1 \\
        M, S, F, s & each timestamp (F) and sensor ID (S) pair is mapped to a value (M) with (s) & a water flow sensor with the ID s1, measures u1 at a time t1 \\
        V, S, b & each sensor (S) is mapped (b) to a node (V) in the knowledge graph & (S1:Sensor)\_(:has\_type)\_(:WaterFlow), a water flow sensor \\
        \bottomrule
    \end{tabular}
\end{table}

Note that the term \textit{semantic association rules} is also used when mining rules from knowledge graphs~\citep{barati2017mining} only, which is a different task than rule learning from sensor data presented in this paper. To the best of our knowledge, there has been no fully DL-based ARM algorithm for learning association rules from concurrent events in IoT data.

\textbf{Our approach.} In contrast to existing work, we utilize both static knowledge graphs and dynamic sensor data that represent IoT data as a whole and propose a novel neurosymbolic ARM approach for learning \textit{semantic} rules from IoT data, for concurrent events. Our approach leads to a more concise set of high-quality association rules that are more generically applicable than sensor-only rules with full coverage over the data. In addition, semantic association rules facilitate domain knowledge integration as domain knowledge can also be represented as semantic rules, e.g., as part of a domain ontology underlying the knowledge graph.

\section{Problem Definition} \label{sec:problem-statement}


This research problem relates to learning association rules from sensor data in IoT systems with semantic properties from a knowledge graph describing the system and its components. 

Given a sensor dataset T with sensors mapped to nodes in knowledge graph G with binding B, produce a set of association rules with clauses based on T and G. Association rules are formal logical formulas in the form of implications, e.g. $X \rightarrow Y$, where $X \rightarrow Y$ is a horn clause with $|Y| = 1$ referring to a single literal and $|X| \geq 1$ referring to a set of literals. $X$ is referred to as the antecedent, and $Y$ is the consequent. A horn clause is defined as \textit{a disjunction of literals with at most one positive literal}. Note that $p \rightarrow q \wedge r$ can be re-written as $p \rightarrow q$ and $p \rightarrow r$, hence $|Y| = 1$.

\textbf{Note} that the T is converted to a set of transactions before the learning process, e.g., by grouping sensor data based on time frames. G is in the form of a directed property graph which contains semantic information of the items in T, e.g., where a sensor is placed, and binding B maps sensors in T to a corresponding node in G, assuming that each sensor has a representation in G. Output rules can express conditions on the sensor measurements and its context.

\subsection{Input}\label{sec:definition-input}

This section presents input notation. To help readers understand easier, Table \ref{tab:input-notation} lists symbols used in the notation, high-level explanations, and examples from the water network domain.

\textbf{Knowledge graph.} The knowledge graph described in this section is a property graph with an ontology or data schema as the underlying structure~\citep{tamavsauskaite2023defining}. We adapt the definition for a \textit{property graph}, given in the next paragraph, from \citep{kg-book}.

\textbf{Property Graph.} Let \textit{Con} be a countably infinite set of constants. A property graph is a tuple $G=(V, E, L, P, U,e,l,p)$, where $V \subseteq Con$ is a set of node IDs, $E \subseteq Con$ is a set of edge IDs, $L \subseteq Con$ is a set of labels, $P \subseteq Con$ is a set of properties, $U \subseteq Con$ is a set of values, $e:E \rightarrow V \times V$ maps an edge ID to a pair of node IDs, $l:V \cup E \rightarrow 2^L$ maps a node or edge ID to a set of labels, and $p:V \cup E \rightarrow 2^{P \times U}$ maps a node or edge ID to a set of property–value pairs.

\textbf{Ontology/Data Schema.} Let $O=(C, R, A, r, a)$ be an ontology or data schema, where $C \subseteq Con$ is a set of classes, $R \subseteq Con$ is a set of relations, $A \subseteq Con$ be a set of properties, $r:R \rightarrow C \times C$ maps a relation to a pair of classes, and $a:C \cup R \rightarrow 2^P$ maps a class or a relation to a set of properties. 

To express that G has O as its underlying structure, we define; i) $L \subseteq C \cup R$, meaning that the labels in G can only be one of the classes or relations defined in O, ii) $P \subseteq A$, meaning that the properties of V and E in G, can only be one of the properties in A.

\textbf{Sensor data.} We define sensor data generically as a tuple $T = (M, S, F, s)$, where $M \subseteq (\mathbb{R} \cup Con)$ is either real numbers representing numerical sensor measurements or constants representing categorical sensor values (states, e.g., a door is open or closed), $S \subseteq Con$ is a set of sensor IDs, $F$ is an ordered numerical sequence of timestamps and $s:(S, F) \rightarrow M$ maps every sensor ID and timestamp to a value. \textbf{Note}, further in this approach, the order of timestamps is considered only to aggregate sensor measurements into transactions (of time frames) to enable generalizable rule learning, since the task is not to learn temporal rules.


\textbf{Binding.} It is a tuple $B = (V, S, b)$, where $V$ is the set of node IDs from G, and $S$ is the set of sensors IDs from T, $b: S \rightarrow V$ maps each sensor ID to a node in G, and $b(S) \subseteq V$ meaning that there is a node ID for each sensor ID, and there can be node IDs for more e.g., instances of classes in C. 

\begin{table}[t]
    \centering
    \caption{Output item forms, explanations, and examples from water network domain.}
    \label{tab:output-examples}
    \begin{tabular}{p{0.2\textwidth}p{0.21\textwidth}p{0.38\textwidth}}
        \toprule
        \textbf{Form} & \textbf{Example} & \textbf{Explanation} \\ 
        \midrule
        $(i' = (p' \# z'))$ & (p1:).length $>$ 100 & A node p1 has length bigger than 100 \\
        $(i' = (m' \# z'))$ & (s1:Sensor).value $<$ 10 & A sensor s1 measures a value smaller than 10 \\
        $(i' = (v'_l = l'))$ & $(p1:Pipe)$ & A node p1 has the label 'Pipe' \\
        $(i' = (e'_l = l'))$ & $(e1:Junction)$ & An edge e1 has the label 'Junction' \\
        $(i' = (v' \rightarrow v'' = e'))$ & $(p1:) \rightarrow (p2:) = (e1:)$ & node p1 is connected to p2 with the edge e1 \\
        \bottomrule
    \end{tabular}
\end{table}

\subsection{Output}\label{definition:output}

The output is a set of rules of the form described below.

Let $I$ be a set of items. We define the following forms for an item, which are basic comparison operations: $\forall i' \in I (((i' = (p' \# z')) \lor (i' = (m' \# z')) \lor (i' = (v'_l = l')) \lor (i' = (e'_l = l')) \lor (i' = (v' \rightarrow v'' = e')) )$, with $p' \in P$, $m' \in M$, $v', v'' \in V$, $e' \in E$, $l', v'_l, e'_l \in L$ where $v'_l$ refers to a label mapped to a node with the ID $v'$, and $e'_l$ refers to a label mapped to an edge with the ID $e'$. $z'$ refers to a value that is either \textit{categorical} or \textit{numerical}, $\#$ refers to one of the comparison operations with a truth value defined below:

\noindent
$\#_{categorical}(p, g) ::= (p = g) | (p \neq g) | (p \in \{g\}) | (p \notin \{g\}) $
\noindent
$\#_{numerical}(p, g) ::= (p = g) | (p \neq g) | (p > g) | (p < g) | (p \leq g) | (p \geq g)$

$X \rightarrow Y$ is an association rule where $(X, Y \subseteq I) \wedge (|Y| = 1)$. This means that items of the rule can only consist of properties of classes or relations defined in the ontology, and the consequent can only have 1 item. Examples and explanations for item forms are given in Table \ref{tab:output-examples}. The item forms consist of comparisons over $m \in M$ or $p \in P$, labels $l \in L$, and whether an edge $e \in E$ exists for a pair of $v \in V$. We call rules in this form \textbf{semantic association rules}.

\section{Semantic Association Rules from IoT Data} \label{sec:methodology}

This section introduces our proposed ARM pipeline for IoT data and an Autoencoder-based Neurosymbolic ARM approach (Aerial) as part of the pipeline. The goal is to learn a concise set of high-quality semantic association rules from sensor data and knowledge graphs with full coverage over the data.

\begin{figure}[t]
    \centering
    \includegraphics[width=0.9\textwidth]{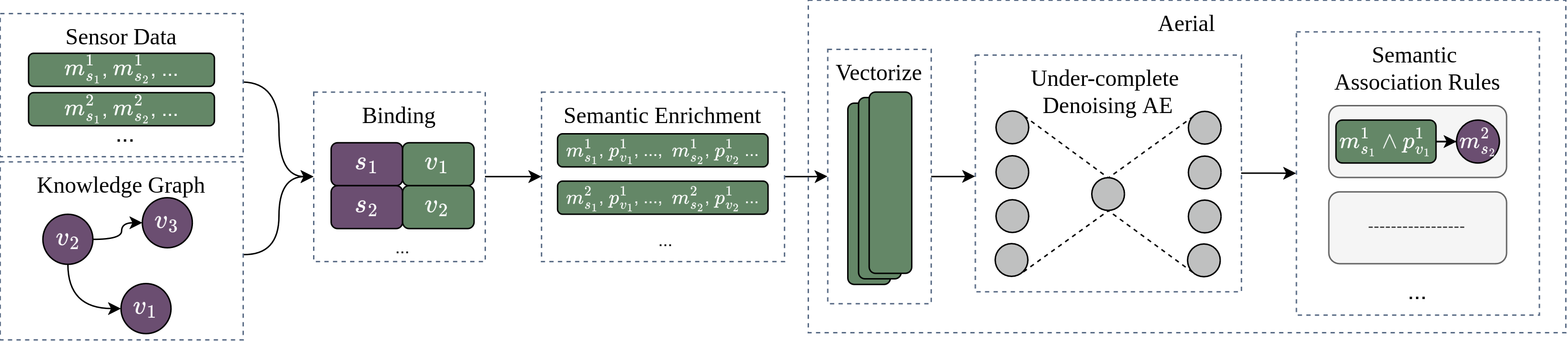}
    \caption{Proposed ARM pipeline for IoT data to learn semantic association rules from sensor data and knowledge graphs.}
    \label{fig:pipeline}
\end{figure}

\subsection{Pipeline}\label{sec:pipeline}

Figure \ref{fig:pipeline} depicts the proposed full pipeline of operations. First, sensor data is aggregated into time frames (e.g., average measurements per minute), hence, forming transactions. Each row in the Sensor Data depiction in Figure \ref{fig:pipeline} refers to a transaction, representing the state of the IoT system at a certain moment in time. Second, binding B is utilized to enrich sensor data with semantics from the knowledge graph. Let $j$ be the number of sensors in $S$, $i$ be the number of semantic property values in $U$ mapped to each $s_{1..j}$, $z$ be the number of classes per input feature for simplicity, and $n$ be the number of transactions. In practice, $i$ and $z$ usually are different per $s_{1..j}$, and property values $p \in U$ can be different per transaction if G changes over time. Property values from neighbors of node $v$ can also be in the transaction set depending on the application. 

Third, in the vectorize step, semantically enriched sensor data is then one-hot encoded and fed into an under-complete denoising Autoencoder~\citep{denoisingautoencoder}. The Autoencoder creates a neural representation of the input data. Our Autoencoder architecture is described in \hyperref[sec:autoencoder-arch]{Autoencoder Architecture} section and the training process is described in \hyperref[sec:training]{Training and Execution} section. Input transactions to the Autoencoder look as follows:

$[\{m1_{s_1}^1, ..., m1_{s_j}^z, p1_{{s_1}_1}^1, ..., p1_{{s_1}_1}^z, ..., p1_{{s_1}_i}^z, ..., p1_{{s_j}_i}^z\}, \\
\indent ... \\ 
\indent \{mn_{s_1}^1, ..., mn_{s_j}^z, pn_{{s_1}_1}^1, ..., pn_{{s_1}_1}^z, ..., pn_{{s_1}_i}^z, ..., pn_{{s_j}_i}^z\}]$

The final step is to extract association rules from a trained Autoencoder which is described in Section \hyperref[sec:rule-extraction-autoencoder]{Rule Extraction from Autoencoders}. Note that some parts of the architecture are kept flexible as they may vary depending on the downstream task that the proposed approach is applied to, such as the type of discretization, sensor data aggregation, encoding, etc.  

\subsection{Autoencoder Architecture}\label{sec:autoencoder-arch}

We employ an under-complete denoising Autoencoder~\citep{denoisingautoencoder} which creates a lower dimensional representation of the noisy variant of its input (encoder) and then reconstructs the noise-free input from the dimensionally reduced version (decoder). In this way, the model learns a neural representation of the input data and becomes more robust to noise. Our under-complete denoising autoencoder has 3 layers for encoding and decoding units. During training, $tanh(z) = \frac{e^z - e^{-z}}{e^z + e^{-z}}$ is preferred in the hidden layers and $softmax(z_i) = \frac{e^{z_i}}{\sum_{j=1}^{n} e^{z_j}}$ preferred at the output layer, as activation functions. The softmax function is applied per category of features so that probabilities per class values are obtained for each category. As the lost function, aggregated binary cross-entropy loss, $BCE\_Loss = \frac{1}{n} \sum_{i=1}^{n} -(y_i log(p_i) + (1-y_i) log(1-p_i))$, is applied to each feature to calculate the loss between Autoencoder reconstruction and the initial noise-free input. The training process is described in Section \hyperref[sec:training]{Training and Execution}.

\subsection{Rule Extraction from Autoencoders} \label{sec:rule-extraction-autoencoder}

The last step of our pipeline is to extract association rules from a trained Autoencoder using \hyperref[alg:aerial]{Algorithm 1}. Aerial is a Neurosymbolic approach to rule mining as it combines a neural network (an Autoencoder) and an algorithm that can extract associations in the form of logical rules from a neural representation of input data created by training the Autoencoder. \textbf{Note} that any other ARM algorithm can be used within the pipeline after the semantic enrichment. 

\textbf{Intuition:} Aerial exploits the reconstruction loss of a trained Autoencoder to learn associations. If reconstruction for an input vector with marked features is more successful than a \textit{similarity threshold} then we say that the marked features imply the successfully reconstructed features. Marking features is done by assigning 1 (100\%) probability to a certain class value for a feature, 0 to the other classes for the same feature, and assigning equal probabilities to the rest of the features in an input vector. 

\begin{wrapfigure}{l}{0.6\textwidth}
    \centering
    \includegraphics[width=0.6\textwidth]{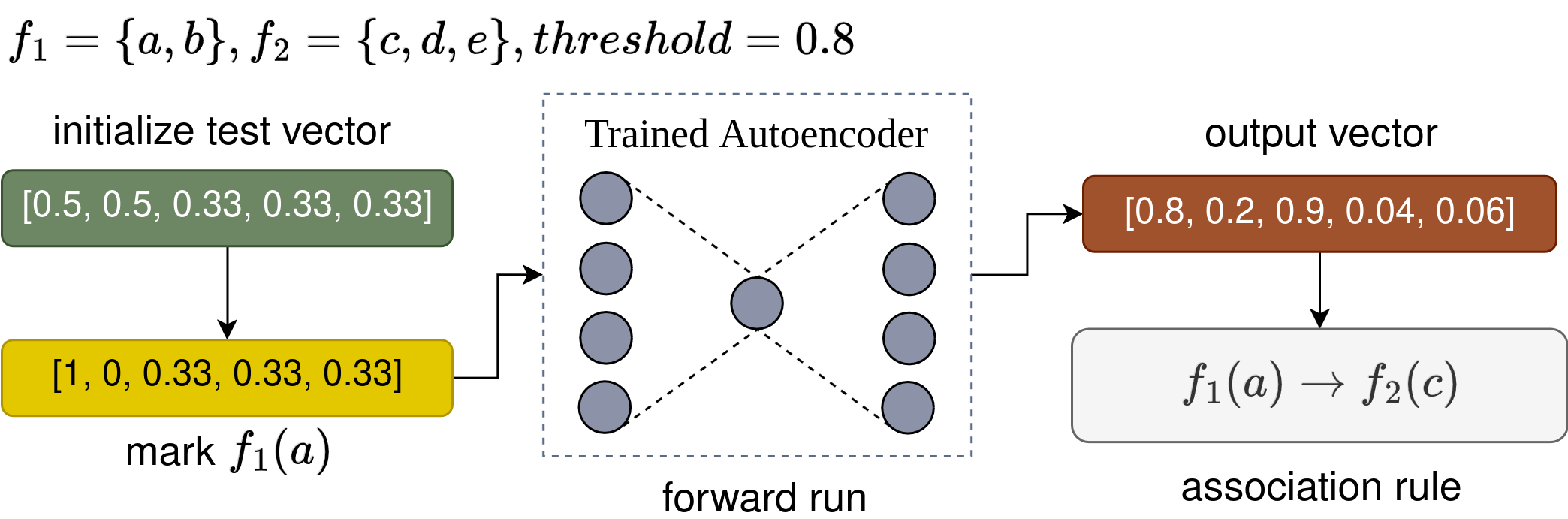}
    \caption{An illustration of association rule extraction from a trained Autoencoder with our Aerial approach.}
    \label{fig:example}
\end{wrapfigure}

\textbf{Example:} Figure \ref{fig:example} depicts an example rule extraction process. Assume that there are only two features in the input vector with 2 and 3 possible class values, namely $f_1 = \{a,b\}$ and $f_2=\{c,d,e\}$. One-hot encoded version of $f_1$ and $f_2$ can be represented with 5 digits. Assume we want to test whether $f_1(a)$ implies a certain class value of $f_2$. Here we do a forward run on the trained Autoencoder with the input vector $[1,0,0.33,0.33,0.33]$ in which $f_1(a)$ is marked with 1, and equal probabilities are given for the values of $f_2$. We call this a $test\_vector$. Assume that the output is $[0.8,0.2,0.9,0.04,0,06]$. The third output digit that corresponds to $f_2(c)$ is bigger than the threshold, 0.8. Therefore, we say that $f_1(a) \rightarrow f_2(c)$.

\textbf{Algorithm:} The rule extraction algorithm is given in Algorithm \ref{alg:aerial}. The parameters are the set of input vectors (\textit{input}), a trained Autoencoder (\textit{ae}), a similarity threshold (\textit{sim\_threshold}), and a maximum number of antecedents (\textit{antecedents}) that the rules will contain. Based on the $antecedents$, in line 3, the algorithm creates combinations of features to be tested ($test\_ftrs)$, for instance, to test whether values of features $f_1$ and $f_2$ are associated with other features, a tuple of $(f_1, f_2)$ is created. Lines 4-13 go through each feature tuple ($ftr\_list$) in the test features and first create an initial test vector with all equal probabilities per feature (line 5). Line 6 marks feature values in the $ftr\_list$ with a probability of 1, and returns a list of test vectors ($test\_vectors$). Lines 7-13 perform a forward run per test vector and; i) check whether output probabilities for the marked features are higher than the given threshold (lines 9-10), ii) find features (other than marked features) that have higher probability than the given threshold, which are added to the rule list as consequences together with the marked features which are the antecedents (lines 11-13). 
The algorithm's \textbf{time complexity} is $O(\binom{f}{a})$, where \textit{f} represents the number of features and \textit{a} denotes the maximum number of antecedents (see the section below for details).

\begin{algorithm}[t]
	\caption{Aerial rule extraction algorithm} 
    \begin{algorithmic}[1]
    \Procedure{ARM}{input, ae, sim\_threshold, antecedent}
        \State rules = []
        \State test\_ftrs = combinations(input.features, antecedent)
        \For {ftr\_list in test\_ftrs}:
            \State init\_vector = equal\_prob\_vector(input.features)
            \State test\_vectors = mark\_ftrs(ftr\_list, init\_vector)
            \For {vector in test\_vectors}:
                \State out\_probs = ae.forward(vector)
                \If {out\_probs(ftr\_list) $<$ sim\_threshold}:
                    \State \textbf{continue} with\_next\_test\_vector 
                \EndIf
                \For {feature in out\_probs - ftr\_list}:
                    \If {feature.prob() $>$ sim\_threshold}:
                        \State rules.append(\{antecedent: ftr\_list, consequent: feature.index\})
                    \EndIf
                \EndFor
            \EndFor
        \EndFor
        \State \Return rules
    \EndProcedure
    \end{algorithmic}
    \label{alg:aerial}
\end{algorithm}

ARM-AE~\citep{berteloot2023association}, another Autoencoder-based ARM method, uses an Autoencoder with equal size layers (no dimensionality reduction), does not distinguish between features (e.g., by applying softmax per features as in our approach) and assumes that input to the trained Autoencoder represents consequent while the output represents an antecedent. We argue that this assumption does not hold and the evaluation of ARM-AE resulted in exceptionally low rule quality. Therefore, we opted not to include it in the core \hyperref[sec:experiments]{Evaluation} section. Please refer to \hyperref[appendix:arm-ae]{Experiment 6 in Appendices} for the evaluation of ARM-AE.

\subsubsection{Time Complexity Analysis of Aerial.}\label{sec:time-complexity-analysis} This section provides a time complexity analysis of our Aerial approach, Algorithm \ref{alg:aerial}, in big O notation. We analyze each line in the algorithm and aggregate the results at the end.

Line 2 initializes the $rules$ array, therefore it is $O(1)$. 


Line 3 is a combination operation over the input features, \textit{input.features}, taken \textit{antecedent} at a time. Let's assume $f$ is the total number of features, and $a$ is the maximum number of antecedents parameter, then the complexity is $O(\binom{f}{a})$. 


Line 4 iterates over the $test\_ftrs$. Therefore, the operations inside the loop are repeated $\binom{f}{a}$ times. 


Line 5 initiates a vector with equal probabilities per feature class values. It is linear over the feature count, $O(f)$. 


Line 6 creates a set of vectors in which class values of the features in $ftr\_list$ are marked with 1. In the worst-case scenario, this step is linear over features when the $ftr\_list$ is equal to all of the features in the input dataset, hence, $O(f)$. 


Assuming that line 6 generated $m$ vectors, line 7 iterates $m$ times over the generated vectors. 


Line 8 performs a forward pass with the given test vector. Since each forward pass performs a $softmax$ operation over the class values of features, this operation is linear over the number of features, $O(f)$, assuming that softmax is performed in $O(1)$.


Lines 9 and 10 perform a comparison operation to check whether probabilities inside the $out\_probs$ array that corresponds to the marked features are higher than a threshold or not. Assuming the worst-case scenario, this operation is repeated for each feature in the input data, $O(f)$. 




Aggregation of the results:
\begin{enumerate}
    \item The outer loop runs $\binom{f}{a}$ times.
    \item For each iteration of the outer loop, lines 5 and 6 create an initial vector with equal probabilities and mark some of the features in $O(f)$ time.
    \item The middle loop (line 7) runs over the $m$ test vectors. A forward pass and the probability check in lines 8-10 are performed in $O(f)$ time.
    \item The inner-most loop (line 11) runs in $O(f)$ time.
\end{enumerate}

Therefore, the complexity is $O(\binom{f}{a}) \times O(f) \times O(f \times m) \times O(f)$. Assuming that $m$ is linear over the number of features $f$, and $\binom{f}{a}$ being the most expensive operation, the time complexity of Algorithm \ref{alg:aerial} is $O(\binom{f}{a})$.

\section{Evaluation} \label{sec:experiments}

\textbf{Two different experimental settings} are used to evaluate the two main contributions of this paper; i) evaluation of utilizing semantics with sensor data for ARM in comparison to ARM on sensor data only, and ii) evaluation of the proposed Aerial approach in comparison to state-of-the-art ARM algorithms.

This section first describes common elements across both settings such as datasets, and then describes setting-specific points including baselines. Additional experiments that are not directly relevant to the two settings are given in Appendix \hyperref[appendix:additional-experiments]{Additional Experiments}.


\textbf{Open source.} The source codes of Aerial, baselines, and knowledge graph construction are written in Python and are available online together with all the datasets: \hyperlink{https://github.com/DiTEC-project/semantic-association-rule-learning}{https://github.com/DiTEC-project/semantic-association-rule-learning}.

\textbf{Hardware.} \label{sec:hardware} All experiments ran on an AMD EPYC 7H12 64-core CPU with 256 GiB memory. No GPUs were used.

\subsection{Setup}\label{sec:exp-setup}

This section describes the \textbf{common} elements for both of the evaluation settings.

\textbf{Datasets.} 3 open-source IoT datasets from two different domains, water networks and energy, are used for all the experiments. A knowledge graph is created per dataset by mapping metadata about each component to domain-specific data structures. \textbf{LeakDB}~\citep{leakdb} is an artificially generated realistic dataset in water distribution networks. It contains sensor data from 96 sensors of various types, and semantic information such as the formation of the network, sensor placement, and properties of components.
\textbf{L-Town}~\citep{ltown} is another dataset in the water distribution networks domain with the same characteristics. It has 118 sensors.
\textbf{LBNL} Fault Detection and Diagnostics Dataset~\citep{lbnl} contains sensor data from 29 sensors and semantics for Heating, Ventilation, and Air Conditioning (HVAC) systems. As semantic properties, it only includes a \textit{type} property.

\textbf{Training and Execution.} \label{sec:training} The Aerial Autoencoder
is trained for each dataset. The training parameters found via grid search are as follows: learning rate is set to $5e^{-3}$, the models are trained for 5 epochs, Adam~\citep{kingma2014adam} optimizer is used for gradient optimization with a weight decay of $2e^{-8}$, and the noise factor for the denoising Autoencoder is $0.5$. All experiments are repeated 20 times over 20 randomly selected sensors for each dataset, and the average results are presented unless otherwise specified. The random selection is done by picking a random sensor node on the knowledge graph, and traversing through the first, second, etc. neighbors until reaching 20 sensor nodes. Equal-frequency discretization~\citep{foorthuis2020impact} with 10 intervals is used for numerical features for the methods that require pre-discretization (Table \ref{tab:comparison}).

\begin{table}[t]
    \centering
    \caption{Overall comparison of evaluated ARM approaches.}
    \begin{tabular}{p{0.2\textwidth}p{0.17\textwidth}p{0.25\textwidth}p{0.24\textwidth}}
        \toprule
        & \textbf{Exhaustive} & \textbf{DL-based} & \textbf{Optimization} \\
        \midrule
        \textbf{Semantic Assoc. Rules} & Supports & Supports & Does not directly support \\
        \midrule
        \textbf{Rule Constraints} & Supports constraints & Supports constraints & Does not support \\
        \midrule
        \textbf{Number of Rules} & Very high & Low with full data coverage & Medium to High \\
        \midrule
        \textbf{Rule Length} & Controllable & Controllable & Uncontrollable \\
        \midrule
        \textbf{Rule Quality} & Controllable & Partially controllable & Partially controllable \\
        \midrule
        \textbf{Discretization} & Required & Required & Not required \\
        \bottomrule
    \end{tabular}
    \label{tab:comparison}
\end{table}

\textbf{Evaluation Metrics.} The most common way of evaluating ARM algorithms is to measure the quality of the rules from different aspects as there is \textbf{no single criterion} that fits all cases. In the evaluation, we used the standard metrics in ARM literature which are support, confidence, data coverage, number of rules, and execution time~\citep{kaushik2023numerical,telikani2020survey}. In addition, we selected Zhang's metric~\citep{zhangsmetric} to evaluate the association strength of the rules, commonly used in many open-source libraries including MLxtend~\citep{raschkas_2018_mlxtend} and NiaARM~\citep{stupan2022niaarm}. The definitions are given below:

\begin{itemize}
        \item \textbf{Support}: Percentage of transactions with a certain item or rule, among all transactions (D): $support(X \rightarrow Y) = \frac{|X \cup Y|}{|D|}$.
        \item \textbf{Confidence}: Conditional probability of a rule, e.g., given the transactions with the antecedent X in, the probability of having the consequent Y in the same transaction set: $confidence(X \rightarrow Y) = \frac{|X \cup Y|}{|X|}$.
        \item \textbf{Rule Coverage}: Percentage of transactions that contains antecedent(s) of a rule: $coverage(X \rightarrow Y) = support(X)$.
        \item \textbf{Data Coverage}: It refers to the percentage of transactions to which the learned rules are applicable. 
        \item \textbf{Zhang's Metric}: This metric also considers the case in which the consequent appears alone in the transaction set, besides their co-occurrence, and therefore measures dissociation as well. A score of $>0$ indicates an association, 0 indicates independence and $<0$ indicates dissociation: $zm(X \rightarrow Y) = \frac{confidence(X \rightarrow Y) - confidence(X' \rightarrow C)}{max(confidence(X \rightarrow Y), confidence(X' \rightarrow Y))}$ in which $X'$ refers to the absent of $X$ in the transaction set.
\end{itemize}

\textbf{Hyperparameters.} There are 2 parameters to our Aerial approach: similarity threshold and number of antecedents. The effect of similarity threshold on rule quality is investigated in \hyperref[exp:similarity-threshold]{Experiment 3}. The effect of the number of antecedents on execution time and the number of rules learned is investigated in \hyperref[exp:2-1]{Experiment 2.1}.

\subsection{Experimental Settings} \label{sec:experimental-settings}

This section describes the two core experimental settings together with baselines in each setting. Please refer to Table \ref{tab:comparison} for baseline methods described in the settings below.

\begin{table}[t]
    \centering
    \caption{Aerial, baselines, and their parameters (Optimization refers to TS-NARM and Exhaustive to Naive SemRL. See \hyperref[appendix:arm-ae]{Experiment 6 in Appendices} for the evaluation of ARM-AE).}
    \vspace{2mm}
    \label{tab:parameters}
    \begin{tabular}{p{0.12\textwidth}p{0.12\textwidth}p{0.57\textwidth}}
        \toprule
        \textbf{Algorithm} & \textbf{Type} & \textbf{Parameters} \\ 
        \midrule
        Aerial & DL-based & antecedents=2, similarity=0.8 \\
        ARM-AE & DL-based & antecedents=2, likeness=0.8 \\
        \midrule
        DE & Optimization & $F = 0.5, CR = 0.9$ \\
        GA & Optimization & $p_m = 0.01, p_c = 0.8$ \\
        PSO & Optimization & $c_1 = 0.1, c_2 = 0.1, w = 0.8$ \\
        LSHADE & Optimization & $NP_{max} = 18 . NP , NP_{min} = 4 . NP, H = 5, p = 0.1, r^{arc} = 2$ \\
        jDE & Optimization & $F^{(0)} = 0.5, CR^{(0)} = 0.9, \tau = 0.1$ \\
        \midrule
        FP-Growth & Exhaustive & \multirow{2}{0.57\textwidth}{(both) antecedents=2, min\_support=(Aerial.rules. avg\_support/2), min\_confidence=0.8.} \\
        HMine & Exhaustive &  \\
        \bottomrule
    \end{tabular}
\end{table}

\subsubsection{\textbf{Setting 1: Semantics vs without Semantics.}} \label{sec:setting1} To show that semantics can enable learning more generically applicable rules, two different ARM algorithms, our Aerial approach and a popular exhaustive method FP-Growth~\citep{han2000mining}, are run with and without semantically enriched sensor data. Two algorithms are used to show that including semantics is beneficial regardless of the ARM method applied. The results are compared based on the number of rules, average rule support, confidence and coverage, and execution time. FP-Growth is implemented using MLxtend~\citep{raschkas_2018_mlxtend}.

\subsubsection{\textbf{Setting 2: Aerial vs state-of-the-art}.} \label{sec:setting2} The goal is to evaluate the proposed Aerial method for IoT data, and the experiments are run on sensor data with semantics. The only existing semantic ARM approach Naive SemRL~\citep{karabulut2023semantic} is chosen as a baseline and executed with the exhaustive FP-Growth (as in the original paper) and HMine algorithms. In addition, the optimization-based NARM method TS-NARM~\citep{fister2023time} with standard confidence metric as optimization goal is run with 5 algorithms (as in the original paper), Differential Evolution (DE)~\citep{differentialevolution}, Particle Swarm Optimization (PSO)~\citep{pso}, Genetic Algorithm (GA)~\citep{geneticalgorithm}, jDE~\citep{jDE}, and LSHADE~\citep{lshade}). TS-NARM is implemented using NiaPy~\citep{vrbanvcivc2018niapy} and NiaARM~\citep{stupan2022niaarm}, and FP-Growth and HMine are implemented using Mlxtend~\citep{raschkas_2018_mlxtend}. All rule quality criteria described earlier are used in the comparison.

ARM-AE~\citep{berteloot2023association}, another Autoencoder-based ARM method, uses an Autoencoder with equal size layers (no dimensionality reduction), does not distinguish between features (e.g., by applying softmax per features as in our approach) and assumes that input to the trained Autoencoder represents consequent while the output represents an antecedent. We argue that this assumption does not hold and the evaluation of ARM-AE resulted in exceptionally low rule quality. Therefore, we opted not to include it in the core \hyperref[sec:experiments]{Evaluation} section. Please refer to \hyperref[appendix:arm-ae]{Experiment 6 in Appendices} for the evaluation of ARM-AE.

\subsubsection{\textbf{Challenges in comparison.}} The distinct nature of different types of algorithms makes comparability a challenge. The exhaustive algorithms can find all rules with a given support and confidence threshold. The execution time of the 5 optimization-based approaches (TS-NARM) is directly controlled by the pre-set maximum evaluation parameter. And running them longer leads to better results up to a certain point (Section \hyperref[sec:experiments-setting2]{Aerial vs state-of-the-art}). The quality of the rules learned by the DL-based ARM approaches depends on the given similarity threshold parameter (or likeness for ARM-AE). Given these differences, we made our best effort to compare algorithms fairly and showed the \textbf{trade-offs under different conditions}. Table \ref{tab:parameters} lists the parameters of each algorithm for both of the settings, unless otherwise specified. For TS-NARM, the population size is set to 200 which represents an initial set of solutions, and the maximum evaluation is set to 50,000 which represents the number of fitness function evaluations before convergence. The parameters of the 5 optimization-based methods, population size, and maximum evaluation count are the same as in the original paper. The antecedent length of both exhaustive and DL-based ARM methods is set to 2 for fairness unless otherwise specified. The minimum support threshold of the exhaustive methods is set to half of the average support of the rules learned by our Aerial method so that both approaches will result in a similar average support value for fairness.

\subsection{Experimental Results}

This section presents the experimental results for both settings.


\subsubsection{\textbf{Setting 1: Semantics vs without Semantics.}}\label{sec:exp-semantic-ar} \hfill

\textbf{Experiment 1.1: Rule Quality.} Table \ref{tab:exp-semantics} shows the results for running Aerial and FP-Growth with (w-s) and without (wo-s) semantic properties. Average support and rule coverage for both algorithms on all datasets increased significantly upon including semantics. The rule count is increased for FP-Growth with semantics, while it decreased with our Aerial approach. The confidence values did not change significantly.


\begin{table}[t]
    \centering
    \caption{Comparison of ARM on sensor data with semantics (w-s, our pipeline) and without (wo-s), showing a significant increase in support and rule coverage with semantics).}
    \vspace{2mm}
    \begin{tabular}{p{0.1\textwidth}p{0.1\textwidth}p{0.1\textwidth}p{0.11\textwidth}p{0.11\textwidth}}
        \toprule
         & \textbf{\# Rules} & \textbf{Support} & \textbf{Rule Cov.} & \textbf{Confidence} \\
        \cmidrule(lr){2-2}
        \cmidrule(lr){3-3}
        \cmidrule(lr){4-4}
        \cmidrule(lr){5-5}
         & w-s $|$ wo-s & w-s $|$ wo-s & w-s $|$ wo-s & w-s $|$ wo-s \\
        \midrule
        \multicolumn{5}{c}{\textbf{LeakDB}} \\
        FP-Growth & 103K $|$ 9K & \textbf{0.41} $|$ 0.19 & \textbf{0.43} $|$ 0.2 & 0.95 $|$ 0.97 \\ 
        Aerial & 554 $|$ 2.5K & \textbf{0.54} $|$ 0.25 & \textbf{0.59} $|$ 0.3 & 0.91 $|$ 0.87 \\
        \multicolumn{5}{c}{\textbf{L-Town}} \\
        FP-Growth & 25K $|$ 5K & \textbf{0.86} $|$ 0.36 & \textbf{0.9} $|$ 0.38 & 0.96 $|$ 0.96 \\  
        Aerial & 1K $|$ 2.5K & \textbf{0.59} $|$ 0.39 & \textbf{0.65} $|$ 0.45 & 0.91 $|$ 0.88 \\
        \multicolumn{5}{c}{\textbf{LBNL}} \\
        FP-Growth & 7K $|$ 2K & \textbf{0.84} $|$ 0.73 & \textbf{0.85} $|$ 0.75 & 0.98 $|$ 0.99 \\ 
        Aerial & 73 $|$ 258 & \textbf{0.74} $|$ 0.65 & \textbf{0.74} $|$ 0.66 & 1.0 $|$ 0.99 \\
        \bottomrule
    \end{tabular}
    \label{tab:exp-semantics}
\end{table}

The results indicate that association rules learned from sensor data and semantics are more generically applicable than rules learned from sensor data only, as the support and rule coverage values are significantly higher. Furthermore, this experiment is repeated with varying numbers of sensors, and the results (\hyperref[appendix:semantics-sensor-number]{Experiment 4 in Appendices}) show that a higher number of sensors results in more generically applicable rules. The comparison of rule count and confidence for different approaches will be investigated in \hyperref[sec:experiments-setting2]{Experimental Setting 2}.

\begin{figure}[b]
    \centering
    \includegraphics[width=0.55\textwidth]{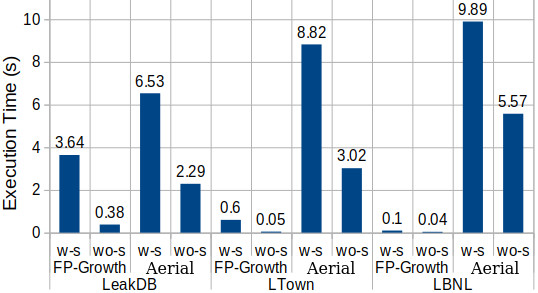}
    \caption{Effect of using semantics (indicated as w-s, and wo-s for without semantics) on execution time.}
    \label{fig:semantics-vs-without}
\end{figure}

\textbf{Experiment 1.2: Execution Time.} \label{exp:1-2} Figure \ref{fig:semantics-vs-without} shows the effect of including semantics in the execution time of FP-Growth and Aerial (training + rule extraction time). The increase in the execution time of FP-Growth is 3-12 times while it is 2-3 times in Aerial and is more stable. However, since the semantic association rules have higher support and data coverage, a smaller number of them can have full data coverage (which is the case for Aerial and will be investigated in \hyperref[sec:experiments-setting2]{Experimental Setting 2}). Therefore, we argue that the increment in the execution time is acceptable. Note that despite FP-Growth running faster with the parameters given in Table ~\ref{tab:parameters}, it is strictly dependent on the preset minimum support threshold value and it runs slower for lower thresholds. This is investigated in \hyperref[sec:experiments-setting2]{Experiment 2.1}. 


\textbf{Illustration.} Table \ref{tab:rule-example} shows two example association rules learned from the LeakDB dataset. The first rule is based on the semantics and sensor data and has higher support and coverage than the second rule, which is only about two specific water flow sensors. 

\subsubsection{\textbf{Setting 2: Aerial vs state-of-the-art}} \label{sec:experiments-setting2}

\begin{table}[t]
    \centering
    \caption{Association rule examples with (top) and without (bottom) semantics learned from LeakDB dataset.}
    \vspace{2mm}
    \label{tab:rule-example}
    \begin{tabular}{p{0.65\textwidth}p{0.08\textwidth}p{0.08\textwidth}}
        \toprule
        \textbf{Association Rule} & \textbf{Support} & \textbf{Coverage} \\
        \midrule
        if a water flow sensor s1 is inside a Pipe with length 843-895, and a water demand sensor s2 inside a Junction measures 13-17, then s1 must measure between 23-31. & 0.5 & 0.54 \\
        \hline
        \rule{0pt}{4mm}
        if the water flow sensor inside Pipe\_28 measures between 23-31, then the water flow sensor inside Pipe\_18 must measure between -767--471. & 0.43 & 0.52 \\
        \bottomrule
    \end{tabular}
\end{table}

\textbf{Experiment 2.1: Execution Time and Number of Rules Analysis.}\label{exp:2-1} This experiment investigates how execution time and the number of rules change for the proposed Aerial approach and baselines depending on their relevant parameters. 

\begin{wrapfigure}{l}{0.55\textwidth}
    \centering
    \includegraphics[width=0.55\textwidth]{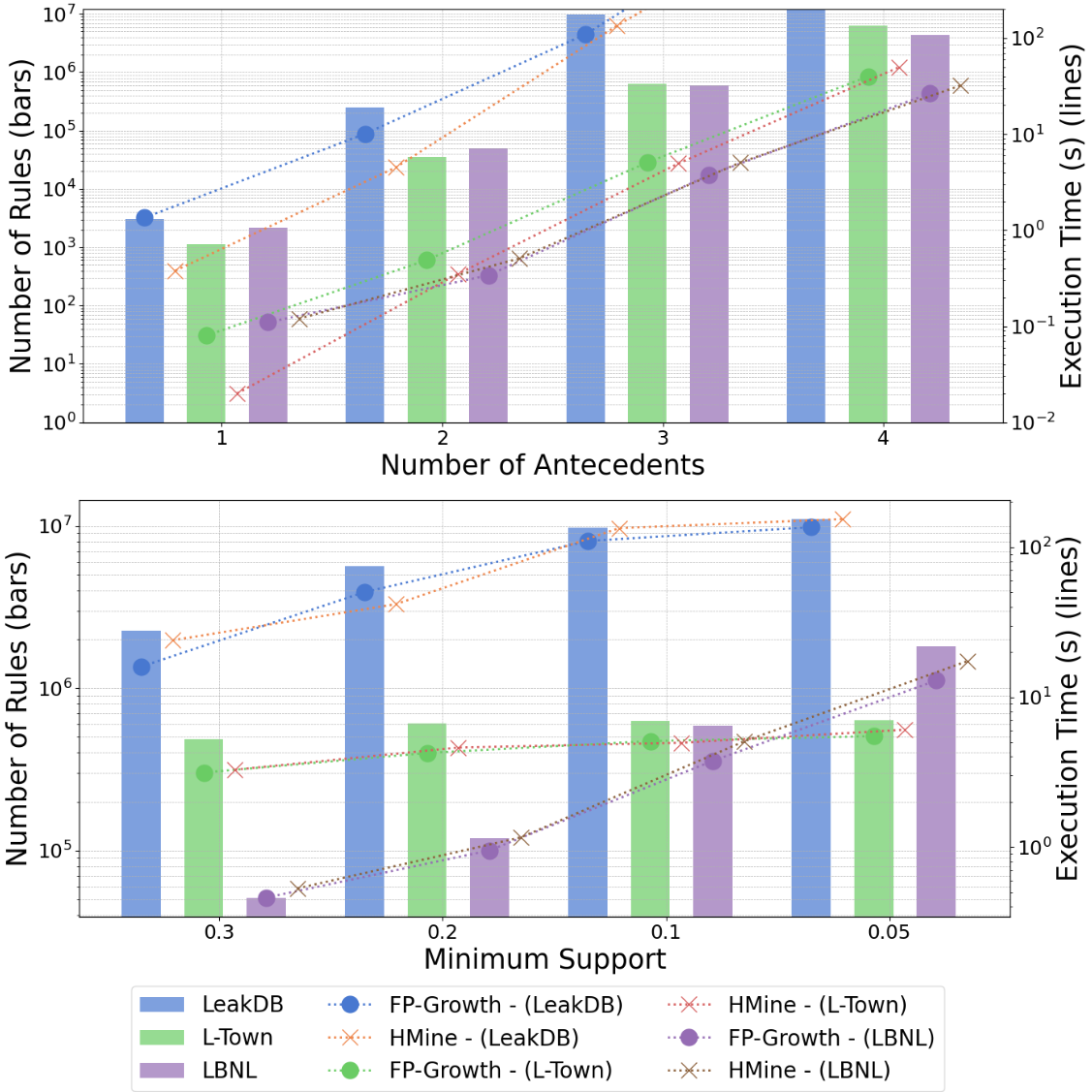}
    \caption{Exhaustive methods have higher execution times (dotted lines) and produce a larger number of rules (bars) as the number of antecedents (top chart, conf=0.8, sup=0.1) increase or min. support threshold (bottom chart, antecedents=3) decrease.}
    \label{fig:exhaustive-changing-support-exec-time-num-rules}
\end{wrapfigure}

The exhaustive methods' execution time and number of rules they mine are strictly dependent on the preset minimum support threshold and the number of antecedents. Figure \ref{fig:exhaustive-changing-support-exec-time-num-rules} shows how the number of rules and execution time change based on antecedents (for 1, 2, 3, and 4 antecedents) and minimum support thresholds (for 0.05, 0.1, 0.2 and 0.3). The results show that the execution time increases as the support threshold decreases and the number of rules increases above 10 million for LeakDB while it reaches 1-2 million for LBNL and L-Town datasets which are highly costly to post-process. Similarly, as the number of antecedents increases the number of rules reaches the levels of millions, while the execution time reaches minutes. The execution did not terminate for the LeakDB dataset when using 4 antecedents after 30 minutes. 

Execution time, number of rules as well as the quality of the rules mined by the optimization-based methods (TS-NARM) strictly depend on the number of evaluations. Table \ref{tab:evaluations-optimization-arm} shows the effect of the maximum evaluations parameter on the execution time, number of rules, and confidence of the rules for the LeakDB dataset (the results are consistent across datasets, see \hyperref[appendix:evaluations-optimization-arm]{Experiment 5 in Appendices}). The results show that longer executions lead to a higher number of rules with higher confidence for all 5 algorithms. 50,000 is chosen as the maximum evaluation for the rule quality experiment (\hyperref[exp:2-2]{Experiment 2.2}) as this is also the case in the original paper.

\begin{table}[t]
    \centering
    \caption{TS-NARM needs long evaluations for good performance (LeakDB, Conf=Confidence). The results are consistent across all datasets (\textbf{\hyperref[appendix:evaluations-optimization-arm]{Experiment 5 in Appendices}}).}
    \vspace{2mm}
    \label{tab:evaluations-optimization-arm}
    \begin{tabular}{p{0.1\textwidth}p{0.09\textwidth}p{0.08\textwidth}p{0.08\textwidth}p{0.1\textwidth}}
    \toprule
    \textbf{Evaluations} & \textbf{Algorithm} & \textbf{\# Rules} & \textbf{Time (s)} & \textbf{Confidence} \\
    \midrule
    \multirow{5}{0.08\textwidth}{10000} & DE & 1388 & 109.24 & 0.69 \\
    & GA & 106 & 120.58 & 0.47 \\
    & PSO & 3281 & 115.39 & 0.81 \\
    & LSHADE & 1786 & 133.01 & 0.77 \\
    & jDE & 1578 & 88.48 & 0.75 \\
    \midrule
    \multirow{5}{0.08\textwidth}{30000} & DE & 6868 & 344.73 & 0.80 \\
    & GA & 472 & 393.73 & 0.40 \\
    & PSO & 10491 & 425.44 & 0.74 \\
    & LSHADE & 9914 & 411.82 & 0.94 \\
    & jDE & 5441 & 300.94 & 0.78 \\
    \midrule
    \multirow{5}{0.08\textwidth}{50000} & DE & 32525 & 782.72 & 0.81 \\
    & GA & 11578 & 650.88 & 0.60 \\
    & PSO & 32502 & 784.96 & 0.84 \\
    & LSHADE & 34887 & 981.07 & 0.99 \\
    & jDE & 24978 & 567.10 & 0.83 \\
    \bottomrule
    \end{tabular}
\end{table}

\begin{figure}[b]
    \centering
    \includegraphics[width=0.48\textwidth]{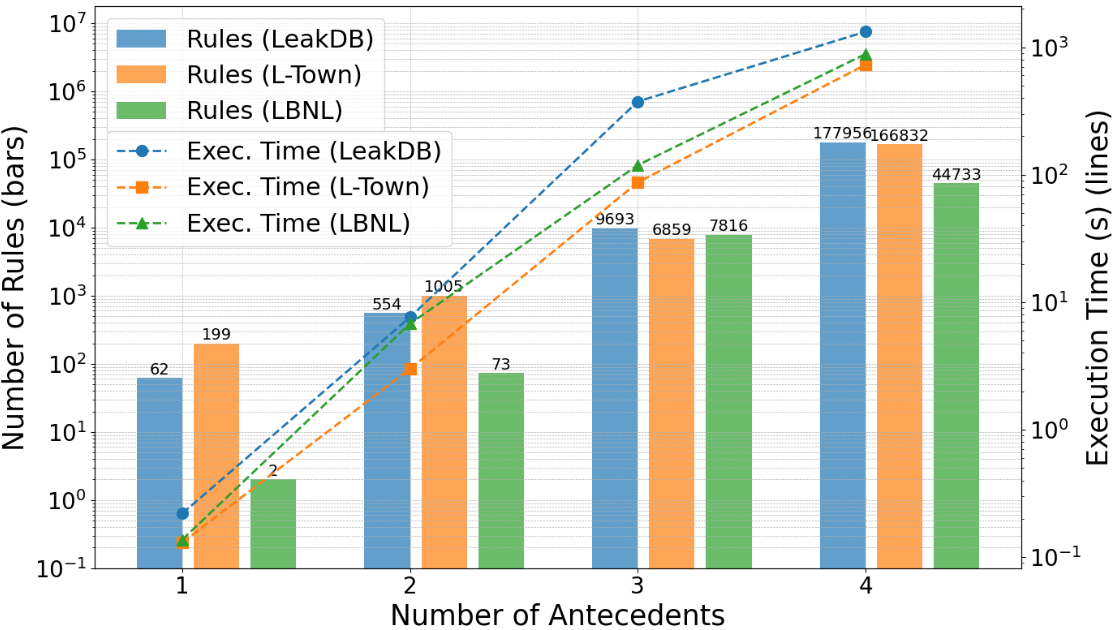}
    \caption{Execution time and the number of rules learned by Aerial depends on the number of antecedents.}
    \label{fig:aerial-changing-antecedent}
\end{figure}

Lastly, the rule extraction time of the proposed Aerial approach is affected by the number of antecedent parameters, as it increases the number of test vectors used in the algorithm. Figure \ref{fig:aerial-changing-antecedent} shows the effect of increasing the number of antecedents on the number of rules and execution time. The number of learned rules is 10-100 times lower than the exhaustive methods. Exhaustive methods run slower on datasets with low support rules, LeakDB (see Tables \ref{tab:exp-semantics} and \ref{tab:settings2-rule-quality}), while running faster on datasets with high support rules, L-Town and LBNL. Both Aerial and exhaustive methods run faster than the optimization-based methods for at least a low-to-medium-size antecedent (1-4).

\textbf{Experiment 2.2: Rule Quality Analysis.} \label{exp:2-2} The goal of this experiment is to assess the quality of rules found by Aerial and baselines, highlighting the trade-offs between algorithms. \textbf{How to read the results?} The evaluation results are shown in Table \ref{tab:settings2-rule-quality} and the highest scores are intentionally \textbf{not} emphasized as ideal rule quality values can vary by task. As an example, high-support rules can be good at discovering trends in the data while low-support rules may be better at detecting anomalies. The focus is on understanding each algorithm's strengths under diverse conditions, therefore, results \textbf{should be interpreted together}.

\begin{table}[t]
    \centering
    \caption{Rule qualities of all algorithms across all datasets (Exhaustive = FP-Growth and HMine.).}
    \begin{tabular}{cccccc}
        \toprule
        \textbf{Algorithm} & \textbf{\shortstack{\# Rules}} & \textbf{Support} & \textbf{Confidence} & \textbf{\shortstack{Data Cov.}} & \textbf{\shortstack{Zhang}} \\
        \midrule
        \multicolumn{6}{c}{\textbf{LeakDB}} \\
        Exhaustive & 103283 & 0.41 & 0.95 & 1.0 & 0.82 \\
        DE & 11841 & 0.19 & 0.77 & 1.0 & 0.24 \\
        GA & 663 & 0.08 & 0.46 & 1.0 & 0.15 \\
        PSO & 12566 & 0.08 & 0.75 & 1.0 & 0.16 \\
        LSHADE & 23605 & 0.4 & 0.98 & 1.0 & 0.41 \\
        jDE & 10270 & 0.25 & 0.77 & 1.0 & 0.29 \\
        Aerial & 554 & 0.54 & 0.91 & 1.0 & 0.9 \\
        \multicolumn{6}{c}{\textbf{L-Town}} \\
        Exhaustive & 25421 & 0.86 & 0.96 & 1.0 & -0.18 \\ 
        DE & 15163 & 0.11 & 0.76 & 1.0 & 0.13 \\
        GA & 1384 & 0.03 & 0.37 & 1.0 & 0.05 \\
        PSO & 15651 & 0.03 & 0.75 & 1.0 & 0.04 \\
        LSHADE & 22825 & 0.39 & 0.96 & 1.0 & 0.39 \\
        jDE & 11255 & 0.19 & 0.78 & 1.0 & 0.21 \\
        Aerial & 1005 & 0.59 & 0.91 & 1.0 & 0.4 \\
        \multicolumn{6}{c}{\textbf{LBNL}} \\
        Exhaustive & 7220 & 0.84 & 0.98 & 1.0 & 0.01 \\
        DE & 17393 & 0.22 & 0.79 & 1.0 & 0.23 \\
        GA & 580 & 0.1 & 0.45 & 1.0 & 0.11 \\
        PSO & 17944 & 0.06 & 0.8 & 1.0 & 0.07 \\
        LSHADE & 30799 & 0.52 & 0.98 & 1.0 & 0.52 \\
        jDE & 15594 & 0.28 & 0.77 & 1.0 & 0.29 \\
        Aerial & 73 & 0.74 & 1.0 & 1.0 & 0.15 \\
        \bottomrule
    \end{tabular}
    \label{tab:settings2-rule-quality}
\end{table}

Aerial was able to find a concise set of rules that have full data coverage with 90\%+ confidence, the highest association strength (Zhang's metric) in the LeakDB and L-Town datasets, and the second highest in the LBNL dataset. The FP-Growth and HMine algorithms yield the same results as they are \textit{Exhaustive}. They have full data coverage, resulted in a high number of rules except for the LBNL dataset, and had very low association strength on L-Town and LBNL. The optimization-based methods had low confidence except for the LSHADE which had a high confidence score on all datasets, the highest association strength among other optimization-based methods, and the highest in LBNL among all algorithms. 

These results show that Aerial was able to find prominent patterns in the datasets that have high association strength and achieved full data coverage with a concise number of rules in comparison to state-of-the-art, which was the initially stated goal. In addition, \hyperref[exp:similarity-threshold]{Experiment 3} shows that higher similarity thresholds in Aerial lead to even higher quality association rules.

\textbf{Experiment 3: Effect of similarity threshold on rule quality in Aerial.}\label{exp:similarity-threshold}
The similarity threshold parameter of our Aerial method affects the quality of the rules learned. This experiment investigates the effect of the similarity threshold parameter of Aerial on all 3 datasets.

\begin{table}[b]
    \centering
    \caption{Aerial learns a more concise set of higher quality rules as the similarity threshold increases.}
    \label{tab:similarity-threshold}
    \begin{tabular}{cccccc}
    \toprule
        \textbf{Threshold} & \textbf{\# Rules} & \textbf{Support} & \textbf{Confidence} & \textbf{Coverage} & \textbf{Zhang} \\
        \midrule
        \multicolumn{6}{c}{\textbf{LeakDB}} \\
        0.9 & 412   & 0.47 & \textbf{0.92} & 1 & \textbf{0.91} \\
        0.8 & 554.4 & \textbf{0.54} & 0.91 & 1 & 0.9  \\
        0.7 & 1845  & 0.3  & 0.88 & 1 & 0.83 \\
        0.6 & 3027  & 0.25 & 0.84 & 1 & 0.79 \\
        0.5 & \textbf{9831}  & 0.28 & 0.73 & 1 & 0.58 \\
        \multicolumn{6}{c}{\textbf{L-Town}} \\
        0.9 & 116 & \textbf{0.7} & \textbf{0.98} & \textbf{1} & 0.06 \\
        0.8 & 1005.2 & 0.59 & 0.91 & \textbf{1} & \textbf{0.4} \\
        0.7 & 1860 & 0.39 & 0.82 & \textbf{1} & 0.33 \\
        0.6 & 3851 & 0.32 & 0.76 & \textbf{1} & 0.32 \\
        0.5 & \textbf{23017} & 0.38 & 0.65 & \textbf{1} & 0.2 \\
        \multicolumn{6}{c}{\textbf{LBNL}} \\ 
        0.9 & 6 & \textbf{0.75} & \textbf{1} & 0.71 & 0 \\
        0.8 & 73 & 0.74 & \textbf{1} & \textbf{1} & \textbf{0.15} \\
        0.7 & 826 & 0.66 & 0.86 & \textbf{1} & 0.13 \\
        0.6 & 1730 & 0.64 & 0.75 & \textbf{1} & 0.08 \\
        0.5 & \textbf{2877} & 0.63 & 0.7 & \textbf{1} & 0.06 \\
    \bottomrule
    \end{tabular}
\end{table}

Table \ref{tab:similarity-threshold} presents the results for all 3 datasets. We observe that as the similarity threshold increases, the number of learned rules decreases, while the average support, confidence, and association strength (Zhang's metric) increase with the exception when the similarity threshold is 0.9. In that case, we observe a decrement in the association strength except in the LeakDB dataset. We argue that this is due to both the relatively low number of rules (6 and 116) learned in comparison to a relatively higher number of rules in LeakDB (412), and LeakDB being a low-support dataset (see Table \ref{tab:exp-semantics}), meaning that the average rule support for association rules in the LeakDB dataset is significantly lower than the other two datasets. 

These results imply that increasing the similarity threshold results in more prominent rules but less in numbers, acting similarly to the minimum confidence threshold of the exhaustive algorithms. 

\subsection{Discussion} \label{sec:discussion}

This section discusses and summarizes the experimental findings.

\textbf{Semantics for generalizability.} The results in \hyperref[sec:exp-semantic-ar]{Experimental Setting 1} showed that learning association rules from both static and dynamic data in IoT systems results in rules that have higher support and data coverage and, therefore, are more generically applicable than rules learned from sensor data only. The experiments also showed that including semantics is beneficial regardless of the ARM approach as the results were similar for both exhaustive FP-Growth and our proposed Aerial approach. 

\textbf{Neurosymbolic methods can help learning a concise set of high-quality rules.} As semantic enrichment of sensor data increases data dimension, current ARM methods result in a higher number of rules which is already identified as a research problem in the ARM literature. As an alternative, our proposed Neurosymbolic Aerial approach can learn a concise number of rules with full data coverage, high confidence, and association strength, which is demonstrated in \hyperref[sec:experiments-setting2]{Experimental Setting 2}. We believe that there is a potential in the direction of neurosymbolic rule learning, and Aerial is a strong initial step.

\textbf{Execution time.} Semantic enrichment increases execution time by 2-3 times for Aerial and 3-12 times for exhaustive methods, as shown in \hyperref[exp:1-2]{Experiment 2.1}. However, semantic association rules have higher support and rule coverage, and a substantially smaller number of them can have full data coverage, therefore we argue that the increment is acceptable. The exhaustive methods perform poorly on low-support (LeakDB) datasets with a low minimum support threshold and also perform poorly with a high number of antecedents as demonstrated in \hyperref[exp:2-1]{Experiment 2.1}. This experiment also showed that Aerial runs faster than the exhaustive methods on low-support datasets and Aerial's execution time does not depend on the datasets' support characteristics. Note that the Aerial can be parallelized and run on GPU (similar to the exhaustive methods). The optimization-based methods' execution time is directly controlled by the preset maximum evaluation parameter. Longer executions are required to obtain higher-quality rules and this also results in a high number of rules, which are costly to process and maintain. Aerial is faster than the optimization-based methods for learning rules with low-to-medium-size antecedents (1 to 4). Note that the number of antecedents for the optimization-based methods can not be controlled. 

\textbf{Variations of Aerial.} Many existing ideas in ARM literature can be integrated into our Aerial approach. For instance, in ARM with item constraints, rules of interest are described using a taxonomy or an ontology and then ARM algorithms focus on those rules only which speeds up the execution and leads to a smaller number of rules~\citep{srikant1997mining,baralis2012generalized}. A similar mechanism can be implemented in Aerial, simply by creating the test vectors in a way that only the items of interest are marked. This will reduce the number of test vectors, and thus reduce the execution time and the number of learned rules. Similarly, top-k rule mining focuses on mining top-k association rules with the highest quality~\citep{fournier2012mining}. An analogous process in Aerial is to find the top-k rules with the highest output probability. As shown in \hyperref[exp:similarity-threshold]{Experiment 3}, higher output probabilities lead to higher quality rules.

\textbf{Scalability.}  Both \hyperref[sec:time-complexity-analysis]{time complexity} and execution time analyses (\hyperref[exp:1-2]{Experiments 1.2 and 2.1}) show that our approach is scalable on large-scale IoT data. The training is linear over the number of features (sensors) and the number of transactions. \hyperref[alg:aerial]{Algorithm 1} is parallelizable as test vectors per feature subsets are created and processed independently. Extrapolating the execution times (training + rule extraction) shown in Figure \ref{fig:aerial-changing-antecedent}, Aerial can scale up to tens of thousands of sensors on a laptop (see \hyperref[sec:hardware]{Hardware}) in a day. 

\section{Conclusion and Future Work} \label{sec:conclusion}

This paper introduced two contributions; i) a novel ARM pipeline for IoT systems, and lii) a Neurosymbolic ARM method (Aerial). In contrast to the state-of-the-art, our pipeline utilizes both dynamic sensor data and static knowledge graphs that describe the metadata of IoT systems. Aerial creates a neural representation of given input data using an Autoencoder and then extracts association rules from the neural representation. The experiments showed that the proposed pipeline can learn rules with 2-3 times higher support and coverage, which are more generically applicable than ARM on sensor data only. Moreover, the experiments further demonstrated that Aerial can learn a more concise set of high-quality association rules than the state-of-the-art with full data coverage. Aerial is also compatible with existing work on addressing the high number of rule problems in the ARM literature.

In future work, we first plan to investigate other neural network architectures for their capabilities of learning associations and develop new methods to extract rules from neural representations created using various architectures. Secondly, we plan to apply our methods to downstream tasks such as leakage detection in water networks, or fault diagnosis in energy systems. 

%
%
%



\textbf{Acknowledgement.} This work has received support from The Dutch Research Council (NWO), in the scope of the Digital Twin for Evolutionary Changes in water networks (DiTEC) project, file number 19454.

\bibliographystyle{unsrtnat}
\bibliography{main}

\appendix

\section{Additional Experiments} \label{appendix:additional-experiments}

This section contains auxiliary experiments that were not included in the core part of the paper. The experimental setups described in Section \hyperref[sec:experimental-settings]{Experimental Settings} are followed for these additional experiments as well unless otherwise specified.

\begin{table}[t]
    \centering
    \caption{Comparison of ARM on sensor data with semantics (indicated as w-s) and without semantics (wo-s) for 10, 15, and 20 sensors.}
    \begin{tabular}{p{0.14\textwidth}p{0.12\textwidth}p{0.1\textwidth}p{0.1\textwidth}p{0.1\textwidth}}
        \toprule
         & \textbf{\# Rules} & \textbf{Support} & \textbf{Coverage} & \textbf{Confidence} \\
        \cmidrule(lr){2-2}
        \cmidrule(lr){3-3}
        \cmidrule(lr){4-4}
        \cmidrule(lr){5-5}
         & w-s $|$ wo-s & w-s $|$ wo-s & w-s $|$ wo-s & w-s $|$ wo-s \\
        \midrule
        \multicolumn{5}{c}{\textbf{LeakDB}} \\
        FP-Growth (10) & 43K $|$ 472 & \textbf{0.23} $|$ 0.22 & \textbf{0.24} $|$ 0.23 & 0.96 $|$ 0.96 \\ 
        FP-Growth (15) & 130K $|$ 7321 & \textbf{0.27} $|$ 0.14 & \textbf{0.28} $|$ 0.15 & 0.96 $|$ 0.97 \\ 
        FP-Growth (20) & 103K $|$ 8974 & \textbf{0.41} $|$ 0.19 & \textbf{0.43} $|$ 0.2 & 0.95 $|$ 0.97 \\ 
        Aerial (10) & 123 $|$ 109 & \textbf{0.31} $|$ 0.24 & \textbf{0.33} $|$ 0.27 & 0.94 $|$ 0.91 \\ 
        Aerial (15) & 547 $|$ 940 & \textbf{0.31} $|$ 0.27 & \textbf{0.37} $|$ 0.31 & 0.88 $|$ 0.89 \\ 
        Aerial (20) & 554 $|$ 2521 & \textbf{0.54} $|$ 0.25 & \textbf{0.59} $|$ 0.3 & 0.91 $|$ 0.87 \\ 
        \multicolumn{5}{c}{\textbf{L-Town}} \\
        FP-Growth (10) & 11489 $|$ 578 & \textbf{0.58} $|$ 0.35 & \textbf{0.62} $|$ 0.37 & 0.94 $|$ 0.95 \\ 
        FP-Growth (15) & 19447 $|$ 2055 & \textbf{0.76} $|$ 0.33 & \textbf{0.8} $|$ 0.35 & 0.95 $|$ 0.95 \\ 
        FP-Growth (20) & 25421 $|$ 5047 & \textbf{0.86} $|$ 0.36 & \textbf{0.9} $|$ 0.38 & 0.96 $|$ 0.96 \\ 
        Aerial (10) & 72 $|$ 381 & \textbf{0.61} $|$ 0.34 & \textbf{0.67} $|$ 0.39 & 0.92 $|$ 0.88 \\ 
        Aerial (15) & 264 $|$ 1300 & \textbf{0.54} $|$ 0.35 & \textbf{0.6} $|$ 0.42 & 0.9 $|$ 0.87 \\ 
        Aerial (20) & 1005 $|$ 2551 & \textbf{0.59} $|$ 0.39 & \textbf{0.65} $|$ 0.45 & 0.91 $|$ 0.88 \\ 
        \multicolumn{5}{c}{\textbf{LBNL}} \\
        FP-Growth (10) & 25 $|$ 764 & \textbf{0.94} $|$ 0.24 & \textbf{0.94} $|$ 0.24 & 1 $|$ 0.99 \\ 
        FP-Growth (15) & 280 $|$ 181 & \textbf{0.75} $|$ 0.35 & \textbf{0.75} $|$ 0.35 & 1 $|$ 0.99 \\ 
        FP-Growth (20) & 7220 $|$ 2883 & \textbf{0.84} $|$ 0.73 & \textbf{0.85} $|$ 0.75 & 0.98 $|$ 0.99 \\ 
        Aerial (10) & 422 $|$ 14 & \textbf{0.73} $|$ 0.28 & \textbf{0.73} $|$ 0.29 & 1 $|$ 0.97 \\ 
        Aerial (15) & 832 $|$ 61 & \textbf{0.78} $|$ 0.42 & \textbf{0.78} $|$ 0.43 & 1 $|$ 0.99 \\ 
        Aerial (20) & 73 $|$ 258 & \textbf{0.74} $|$ 0.65 & \textbf{0.74} $|$ 0.66 & 1 $|$ 0.99 \\ 
        \bottomrule
    \end{tabular}
    \label{tab:exp-semantics-with-varying-sensors}
\end{table}

\textbf{Experiment 4: Effect of sensor count on rule generalizability.}\label{appendix:semantics-sensor-number} This experiment follows \hyperref[sec:setting1]{Experimental Setting 1} and investigates whether the effect of semantic enrichment of the sensor data is dependent on the number of sensors in terms of the generalizability of the rules learned. Note that we define the generalizability of rules as having high support and high coverage over the data. This is an extension of the Experiments in Section \hyperref[sec:exp-semantic-ar]{Semantics vs without Semantics}. 

Table \ref{tab:exp-semantics-with-varying-sensors} shows the average rule count, support, rule coverage, and confidence of the rules mined by FP-Growth and our Aerial algorithms with varying numbers of sensors (10, 15, and 20) with (w-s) and without (wo-s) the semantic enrichment. On all 3 datasets, regardless of the number of sensors used, the average support and coverage of the rules increased upon semantic enrichment of the sensor data. This is consistent with the results presented in \hyperref[sec:exp-semantic-ar]{Semantics vs without Semantics} section. In addition, the results show that increasing the number of sensors leads to even higher support and rule coverage values on average. The FP-Growth algorithm mined significantly more rules upon semantic enrichment across all datasets, while the number of learned rules decreased for our Aerial approach after semantic enrichment. We argue that due to the static semantic properties in the knowledge graph, the FP-Growth generates a high number of association rules in between those static properties, while this is not the case for Aerial. Lastly, the confidence values did not change significantly. 

\begin{table}[b]
    \centering
    \caption{Evaluation of ARM-AE on all 3 datasets for experimental setting 2.}
    \label{tab:arm-ae}
    \begin{tabular}{cccccc}
        \toprule
        \textbf{Dataset} & \textbf{\# Rules} & \textbf{Support} & \textbf{Confidence} & \textbf{Coverage} & \textbf{Zhang} \\
        \midrule
        LeakDB & 4400 & 0.08 & 0.13 & 0.05 & -0.79 \\
        L-Town & 5600 & 0.08 & 0.11 & 0.1 & -0.89 \\
        LBNL & 3440 & 0.36 & 0.46 & 0.08 & -0.41 \\
        \bottomrule
    \end{tabular}
\end{table}

\textbf{Experiment 5: Effect of maximum evaluations on the execution time and rule quality of optimization-based ARM.}\label{appendix:evaluations-optimization-arm} This section contains the experiments for evaluating the effect of maximum evaluation parameters of the optimization-based methods (TS-NARM) on execution time and rule quality. The experiment results for the LeakDB dataset are already given in \hyperref[exp:2-1]{Experiment 2.1}. Therefore, this section only contains the results for the L-Town and the LBNL datasets and the results are consistent across all datasets.

\begin{table}[t]
    \centering
    \caption{TS-NARM needs high numbers of evaluations for good performance. The table on the left presents the results for L-Town, and the right table presents the results for the LBNL dataset.}
    \label{tab:ltown-lbnl-max-evaluations}
    \begin{minipage}{0.48\textwidth}
        \begin{tabular}{p{0.1\textwidth}p{0.17\textwidth}p{0.15\textwidth}p{0.16\textwidth}p{0.1\textwidth}}
            \toprule
            \textbf{Evals.} & \textbf{Algorithm} & \textbf{\# Rules} & \textbf{Time (s)} & \textbf{Conf.} \\
            \midrule
                \multirow{5}{0.08\textwidth}{1000} & DE & 55.5 & 5.82 & 0.39 \\
                & GA & 46 & 5.53 & 0.26 \\
                & PSO & 67 & 6.52 & 0.4 \\
                & LSHADE & 76 & 7.08 & 0.36 \\
                & jDE & 76 & 3.81 & 0.5 \\
                \midrule
                \multirow{5}{0.08\textwidth}{10000} & DE & 2595 & 158.4 & 0.68 \\
                & GA &  270 & 127.64 & 0.38 \\
                & PSO & 2215.5 & 148.22 & 0.58 \\
                & LSHADE & 2369 & 131.73 & 0.75 \\
                & jDE & 2038.5 & 110.89 & 0.75 \\
                \midrule
                \multirow{5}{0.08\textwidth}{30000} & DE & 7686.5 & 610.8 & 0.75 \\
                & GA & 823.5 & 576.75 & 0.36 \\
                & PSO & 11026.5 & 572.44 & 0.82 \\
                & LSHADE & 12071.5 & 616.04 & 0.94 \\
                & jDE & 6297.5 & 403.73 & 0.76 \\
                \midrule
                \multirow{5}{0.08\textwidth}{50000} & DE & 31673.6 & 778.46 & 0.81 \\
                & GA & 11239 & 591.89 & 0.51 \\
                & PSO & 30570.2 & 828.5 & 0.75 \\
                & LSHADE & 35559.4 & 871.7 & \textbf{0.98} \\
                & jDE & 24245 & 433.69 & 0.78 \\
            \bottomrule
            \end{tabular}
    \end{minipage}
    \hfill
    \begin{minipage}{0.48\textwidth}
        \begin{tabular}{p{0.1\textwidth}p{0.17\textwidth}p{0.15\textwidth}p{0.16\textwidth}p{0.1\textwidth}}
        \toprule
            \textbf{Evals.} & \textbf{Algorithm} & \textbf{\# Rules} & \textbf{Time (s)} & \textbf{Conf.} \\
        \midrule
            \multirow{5}{0.08\textwidth}{1000} & DE & 90 & 6.91 & 0.49 \\
            & GA & 58.5 & 5.26 & 0.51 \\
            & PSO & 104.5 & 6.93 & 0.48 \\
            & LSHADE & 132.5 & 6.15 & 0.47 \\
            & jDE & 172.5 & 4.83 & 0.65 \\
            \midrule
            \multirow{5}{0.08\textwidth}{10000} & DE & 3037.5 & 115.96 & 0.72 \\
            & GA  & 370.5 & 107.77 & 0.51 \\
            & PSO & 2825.5 & 108.82 & 0.78 \\
            & LSHADE & 3372.5 & 95.99 & 0.82 \\
            & jDE & 2919 & 51.2 & 0.73 \\
            \midrule
            \multirow{5}{0.08\textwidth}{30000} & DE & 9751 & 170.16 & 0.74 \\
            & GA & 419 & 185.84 & 0.5 \\
            & PSO & 8933 & 188.72 & 0.96 \\
            & LSHADE & 17624 & 182.64 & \textbf{0.97} \\
            & jDE & 7958 & 111.05 & 0.77 \\
            \midrule
            \multirow{5}{0.08\textwidth}{50000} & DE & 27778.8 & 479.6 & 0.77 \\
            & GA & 7945.4 & 501.37 & 0.47 \\
            & PSO & 25453.8 & 530.74 & 0.79 \\
            & LSHADE & 26864.4 & 787.79 & \textbf{0.97} \\
            & jDE & 20243.2 & 421.82 & 0.77 \\
        \bottomrule
        \end{tabular}
    \end{minipage}
\end{table}

Table \ref{tab:ltown-lbnl-max-evaluations} shows the results for the L-Town and LBNL datasets respectively. Similar to the results for the LeakDB dataset, as the maximum number of evaluation parameters increases, the number of rules, execution time as well as average confidence of the rules increase. The increment in the confidence values decreases as the maximum evaluations increase. These experiments show that optimization-based methods require longer execution times in order to obtain higher-quality rules.

\textbf{Experiment 6: Extracting Association Rules with ARM-AE.}\label{appendix:arm-ae} As mentioned in Section \hyperref[sec:rule-extraction-autoencoder]{Rule Extraction from Autoencoders}, ARM-AE \citep{berteloot2023association} resulted in exceptionally low rule quality and therefore we opted not to include it in the core evaluation. In the original paper, ARM-AE is tested on categorical tabular data only, and to the best of our knowledge, ARM-AE is the only fully DL-based ARM approach, besides our approach, at the time of writing this paper. Note that there are DL-based approaches to sequential ARM~\citep{he2023gan}, however, that is a different task than the one we tackle in this paper. We adapted ARM-AE to work with sensor data as part of our pipeline and used it as a baseline for \hyperref[sec:experiments-setting2]{Experimental Setting 2}. It expects a number of antecedents, a number of rules per consequent, and a likeness (similarity threshold) parameter. The number of antecedents is set to 2, number of rules per consequent is set to the number of rules learned by our Aerial approach divided by the number of features (of the dataset subject to evaluation), and the likeness is set to 80\%, similar to our approach for fairness. 

The evaluation results, given in Table \ref{tab:arm-ae}, show that ARM-AE resulted in exceptionally low rule quality values on all 3 datasets. Therefore, the results were not included in the core part of the paper, however, for the purpose of having complete novel baselines, we included it in this section.

\end{document}